\pdfoutput=1

\documentclass[11pt]{article}

\usepackage[preprint]{acl}
\usepackage{booktabs}
\usepackage{subcaption} 
\usepackage{times}
\usepackage{latexsym}
\usepackage{amsfonts}
\usepackage[compact]{titlesec} 
\usepackage[T1]{fontenc}

\usepackage[utf8]{inputenc}

\usepackage{microtype}

\usepackage{inconsolata}

\usepackage{graphicx}
\usepackage{enumitem}

\hypersetup{
    colorlinks=true,
    linkcolor=blue,
    filecolor=magenta,      
    urlcolor=cyan,
    pdftitle={Overleaf Example},
    pdfpagemode=FullScreen,
    }

\usepackage{amsmath}
\usepackage{cleveref}
\newtheorem{theorem}{Theorem}[section]

\newtheorem{definition}{Definition}[section]

\newcommand{\R}{\mathbb{R}}

\usepackage[normalem]{ulem}

\newcommand{\ignore}[1]{}

\def\p{{\hat{p}}}

\usepackage{tabularx}
\newcolumntype{Y}{>{\centering\arraybackslash}X}
\usepackage{multirow}
\usepackage{bigstrut}

\title{SoS1: O1 and R1-Like Reasoning LLMs are Sum-of-Square Solvers}

\author{
  Kechen Li$^{1}$ \quad Wenqi Zhu$^{3}$ \quad Coralia Cartis$^{3}$ \quad
  Tianbo Ji$^{2}$ \quad Shiwei Liu$^{3}$ \\
  $^{1}$Nanjing University of Aeronautics and Astronautics \\
  $^{2}$School of Transportation and Civil Engineering, Nantong University \\
    $^{3}$Mathematical Institute, University of Oxford \\
      \texttt{kechenli@nuaa.edu.cn},
      \texttt{jitianbo@ntu.edu.cn} \\
   \texttt{\{wenqi.zhu, cartis, shiwei.liu\}@maths.ox.ac.uk} \\ 
   \\ 
}

\begin{document}
\maketitle
\begin{abstract}
Large Language Models (LLMs) have achieved human-level proficiency across diverse tasks,  but their ability to perform rigorous mathematical problem solving remains an open challenge. In this work, we investigate a fundamental yet computationally intractable problem: determining whether a given multivariate polynomial is nonnegative. This problem, closely related to Hilbert’s Seventeenth Problem, plays a crucial role in global polynomial optimization and has applications in various fields. First, we introduce \textbf{SoS-1K}, a meticulously curated dataset of approximately 1,000 polynomials, along with expert-designed reasoning instructions based on five progressively challenging criteria. Evaluating multiple state-of-the-art LLMs, we find that without structured guidance, all models perform only slightly above the random guess baseline (50\%). However, high-quality reasoning instructions significantly improve accuracy—boosting performance up to 81\%. Furthermore, our 7B model, SoS-7B, fine-tuned on SoS-1K for just 4 hours, outperforms the 671B DeepSeek-V3 and GPT-4o-mini in accuracy while only requiring 1.8\% and 5\% of the computation time needed for letters, respectively. Our findings highlight the potential of LLMs to push the boundaries of mathematical reasoning and tackle NP-hard problems. Code is available at \url{https://github.com/Joe-2002/SoS1}. 
\end{abstract}

\section{Introduction}

With Large Language Models (LLMs) reaching human-level proficiency across a diverse range of tasks \citep{brown2020language,singhal2023large,cai2023large,yoshikawa2023large}, their ability to reason has emerged as a central topic of interest \citep{wei2022chain,huang2022towards}. 
Among these, mathematical reasoning stands out as one of the most rigorous and demanding \cite{kant1908critique,hendrycks2021measuring,ahn2024large,liu2024deepseek}. As a result, the ability of LLMs to solve research-level mathematical problems is not only a critical benchmark for evaluating their reasoning capabilities but also has the potential to transform mathematical research and practice. 

Demonstrated by the success of OpenAI o1 \citep{o1} and DeepSeek-R1 \citep{guo2025deepseek}, test-time scaling has emerged as a promising technique for enhancing LLMs' performance in mathematical reasoning \citep{snell2024scaling,welleck2024decoding}. This approach involves prompting LLMs to generate more reasoning steps, either sequentially \citep{snell2024scalingllmtesttimecompute,hou2025advancinglanguagemodelreasoning,lee2025evolvingdeeperllmthinking}  or in parallel \citep{brown2024large,xin2024deepseek}, to increase the accuracy of their final answers. However, the community's primary focus has been limited to relatively simple levels of mathematics, ranging from high school and Olympiad-level problems to early undergraduate topics \citep{aime,hendrycks2021measuring}. Whether the promise of test-time scaling extends to research-level mathematics remains an open question.

\looseness=-11 In this paper, we investigate a fundamental yet formally well-posed problem in mathematics: determining whether a given multivariate polynomial is nonnegative. This question is closely related to Hilbert’s Seventeenth Problem, which was posed by David Hilbert in 1900 as part of his famous 23 problems presented at the International Congress of Mathematicians (ICM) \cite{hilbert1893ternare}, and it remains central to global polynomial optimization. Many key challenges in applied and computational mathematics can be reframed as deciding the nonnegativity of certain polynomials including control theory \cite{parrilo2000structured}, quantum computation \cite{doherty2002distinguishing}, polynomial games \cite{gvozdenovic2007semidefinite}, tensor methods \cite{zhu2024global, ahmadi2023higher} and combinatorial optimization \cite{gvozdenovic2007semidefinite}. 

\begin{figure*}[h]
    \centering
    \includegraphics[width=0.95\linewidth]{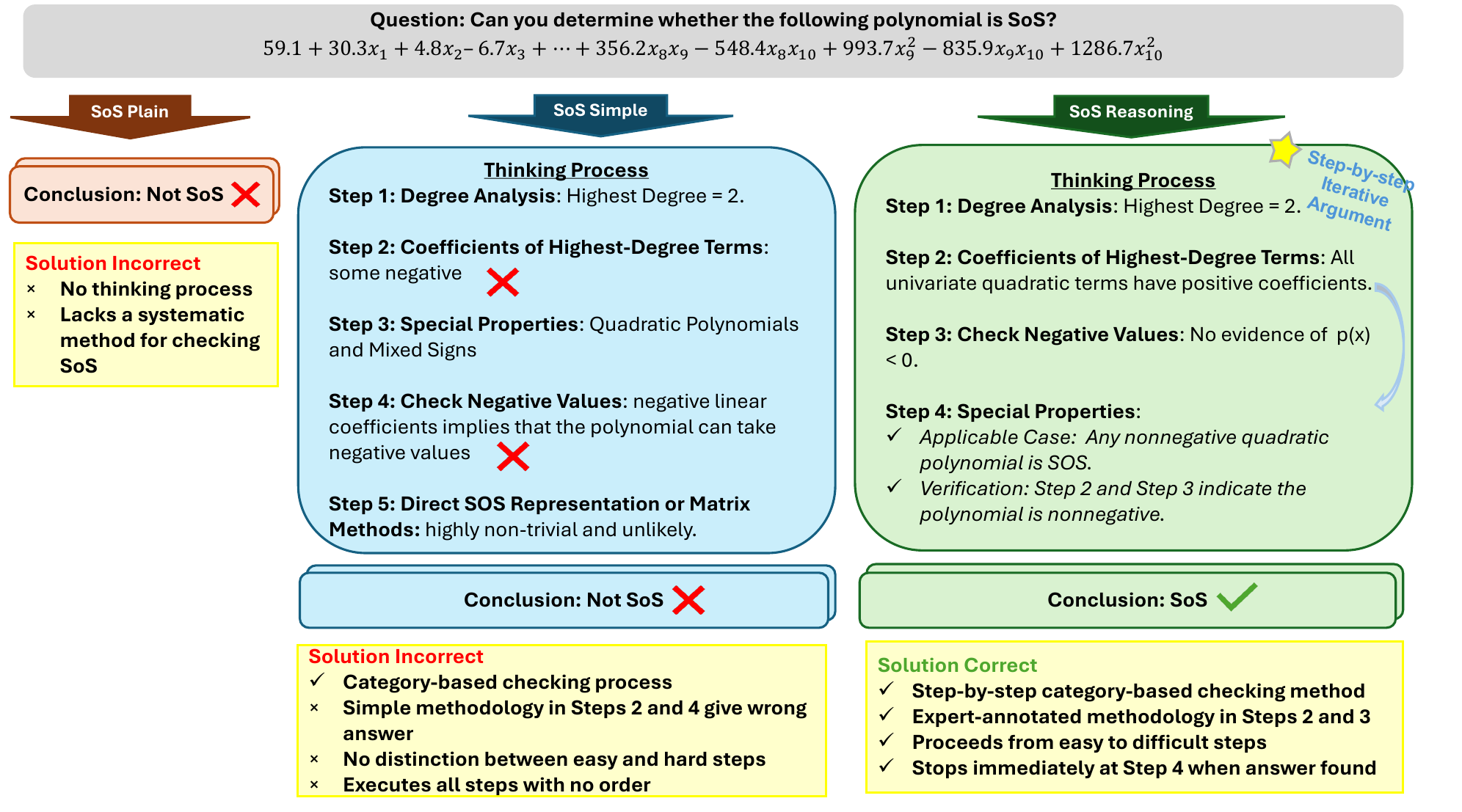}
    \caption{Demonstration of SoS Plain (left), SoS Simple (mid), and SoS Reasoning (right). }
    \label{fig:enter-label}
\end{figure*}

Testing whether a general polynomial is nonnegative is provably NP-hard, even for polynomials of relatively low degrees or with a small number of variables. People usually seek special cases of polynomials where the challenging nonnegativity constraints can be replaced with more manageable conditions. For instance, the sum of squares (SoS) condition, a mathematical technique in polynomial optimization where a polynomial is expressed as a sum of squared polynomials, provides a sufficient criterion for polynomial nonnegativity. Classical solvers, such as \texttt{SoSTOOLS}, \texttt{YALMIP}, and \texttt{Gloptipoly} have been developed to verify these SoS conditions \cite{prajna2002introducing}. However, a significant limitation of these approaches lies in the typically large size of the resulting semidefinite programming (SDP) problem. Specifically, for a polynomial with  $n$ variables and a degree of $2d$, the SDP's dimension is given by $N = \binom{n+2d}{2d}$, making it challenging to scale this approach to larger problems. 

\looseness=-1 To evaluate if state-of-the-art (SOTA) reasoning LLMs like Openai o1 and DeepSeek-R1 can solve large-scale SoS programming problems, we introduce \textbf{SoS-1K}, a meticulously curated dataset of approximately 1,000 polynomials—along with expert-designed, SoS-specialized, reasoning-guiding instructions based on five progressively challenging criteria: polynomial degree, nonnegativity of the leading search direction, identification of special structures, assessment of square-form expressions, and matrix decomposition into the quadratic form of monomials. Our comprehensive evaluation of multiple
SOTA LLMs, including DeepSeek-R1, DeepSeek-V3, GPT-4o, OpenAI o1-mini, Qwen2.5 series, and QwQ-32B-Preview, demonstrate the following interesting findings: 

\begin{itemize}[leftmargin=*]
    \item When presented with a plain question, all SOTA LLMs bluntly fail to solve SoS with most models achieving only around 60\% accuracy, just slightly above the random guess baseline of 50\%.\footnote{Since SoS is a binary classification problem, random guessing yields 50\% accuracy.}

    \item When prompted with high-quality reasoning traces, we consistently observe a significant accuracy boost across all models, up to 21\%. And models perform better with higher-quality reasoning traces.
    
    \item Reasoning-focused LLMs generally outperform general-purpose LLMs, regardless of prompt quality.

    \item Higher-capacity models require fewer thinking tokens to make correct predictions, while lower-capacity models need more reasoning steps to achieve their optimal performance.

\end{itemize}

We further demonstrate that supervised fine-tuning (SFT) of a pre-trained 7B model on SoS-1K for just 4 hours using 2 A100 GPUs significantly improves accuracy from 54\% to 70\% with significantly faster response times. Specifically, SoS-7B requires only 1.8\% and 5\% of the computation time needed for DeepSeek-V3 and GPT-4o-mini, respectively. The resulting SoS-7B surpasses much larger models, including the 671B DeepSeek-V3 and GPT-4o-mini. More interestingly, when prompted with high-quality reasoning, the models demonstrate an understanding of research-level questions. For instance, Qwen2.5-14B-Instruct-1M leverages the Motzkin polynomial to generate new, previously unseen counterexamples to Hilbert’s 17th problem \cite{motzkin1967arithmetic}. Such examples are nontrivial, as the first counterexample to Hilbert’s 17th problem was discovered 27 years after Hilbert originally posed it \cite{hilbert1893ternare}. These findings suggest that LLMs exhibit reasoning patterns, expanding the boundaries of solving NP-hard problems. 

Our work serves as a pilot study on leveraging reasoning LLMs for solving SoS problems, paving the way for tackling large-scale research-level questions in mathematics using AI.

\subsection{Related Work}

The study of SoS and nonnegative polynomials has a rich history spanning over 120 years, with numerous scholars contributing to this field. Classical methods for characterizing an SoS polynomial involve expressing the polynomial as a quadratic form of monomials and then reformulating the problem as a semidefinite program (SDP) \cite{lasserre2000optimisation, lasserre2001global}. Classic techniques, such as the Lasserre hierarchy, date back to 2001 \cite{lasserre2001global}. The development of numerical solvers began in 2009, with tools like \texttt{GloptiPoly} and \texttt{SoSTOOLS} introduced in \cite{henrion2009gloptipoly, papachristodoulou2013SoStools}. More details on the literature review for SoS can be found in \Cref{appendix literature review for SoS}.

The first attempt to tackle SoS-related mathematical difficulty via AI was presented in \cite{alfarano2024global}. The authors trained Transformers to address a long-standing open problem in mathematics: discovering a Lyapunov function that guarantees the global stability of a dynamical system. The global stability of polynomial Lyapunov systems is closely tied to SoS framework. Their approach was tested on relatively small systems (with at most five equations for polynomial systems) and demonstrated promising results compared to state-of-the-art conventional solvers.

\vspace{-0.3em}
\section{SoS-1K Dataset} 
\label{sec:SoS-1k}
\label{sec Reasoning data curation to create SoS1K}
In this section, we first provide the definition of SoS polynomials.  We describe the expert-annotated reasoning traces based on five criteria of increasing difficulty in \Cref{sec cot for SoS1k},  which lead to the generation of the SoS-1K dataset, detailed in \Cref{sec test set}.

\begin{definition} \textbf{SoS polynomial \cite{ahmadi2023higher,ahmadi2013complete, kojima2003sums}}
    A $2d$-degree multivariate polynomial $q(\textbf{x}): \R^n \rightarrow \R$ where $x = [x_1, \dotsc, x_n] \in \R^n$ is a sum of squares (SoS) if there exist polynomials $\Tilde{q}_1, \dotsc, \Tilde{q}_r: \R^n \rightarrow \R$, for some  $r\in \mathbb{N}$, such that $q(\textbf{x}) =\sum_{j=1}^r \Tilde{q}_j(\textbf{x})^2$ for all $\textbf{x} \in \R^n$ \cite[Def. 1]{ahmadi2023higher}. 
    \label{def SoS polynomial}
\end{definition}

\noindent 
Further details on the theory for SoS polynomials can be found in \Cref{appendix theorem for SoS}.

\subsection{Expert-Designed Reasoning Instructions for SoS}
\label{sec cot for SoS1k}

Instructions play a crucial role in guiding LLMs toward better reasoning and problem-solving \citep{zhang2023instruction}. The way an LLM processes a problem can be significantly improved with carefully crafted instructions that provide structure, constraints, and logical flow. 

To evaluate the capacity of SOTA LLMs on SoS reasoning, we create three sets of reasoning-guiding instructions with increasing quality that can be applied across multiple SoS problem types: (1) plain question (SoS Plain); (2) simple SoS instruction (SoS Simple); (3) reasoning-guiding SoS instruction (SoS Reasoning). 

\textbf{SoS Plain} simply asks LLMs: ``Please analyze if this polynomial can be expressed as a Sum of Squares (SOS)''.

\textbf{SoS Simple} classifies SoS polynomials into five distinct groups, each defined by a concise, one-line criterion. The full instruction set for SoS Simple contains 78 words and 647 characters, with complete details provided in \Cref{SoS Simple}.

\textbf{SoS Reasoning} is a structured five-step framework designed to identify SoS polynomials. Unlike SoS Simple, which provides only basic classification criteria, SoS Reasoning encourages the model follow a step by step mathematical  verification process. The framework introduces progressively more detailed reasoning steps to guide the model in verifying whether a polynomial is SoS.  Specifically, we provide a logical reasoning trace based on proofs and theorems, offering necessary and sufficient conditions for identifying SoS polynomials. A large number of positive and negative examples accompany each set of theorems, helping the model recognize special structures, symmetries, and mathematical forms inherent to SoS polynomials. Additionally, SoS Reasoning introduces intermediate steps and incorporates key and challenging reasoning processes, such as the $\mathbf{Q}$ matrix and the squared form $p_s$, enabling deeper reasoning and iterative refinement.

Below is an illustration of SoS Reasoning. The full version is provided in \Cref{SoS Reasoning}. 

\textbf{Step 1. Check the Degree:} An SoS polynomial must have an even highest degree.  

\textbf{Step 2. Check for Non-negativity:} SoS polynomials are nonnegative for all real inputs. We verify this by examining the constant term, the coefficients of the leading term, and performing a grid-based numerical check.  

\textbf{Step 3. Check for Well-known Special Cases:} Any nonnegative quadratic polynomial and any nonnegative quartic polynomial in one or two variables is SoS.  

\textbf{Step 4. Check for Square Form:} By \Cref{def SoS polynomial}, an SoS polynomial can be expressed as:  
$
    p_s(\textbf{x}) = \sum_i q_i(\textbf{x})^2,
$
where each \( q_i(\textbf{x}) \) is a polynomial.

\textbf{Step 5. Check for Matrix Decomposition:} Based on \Cref{thm SoS alternative}, we express the polynomial as  
$
    p(\textbf{x}) = {\textbf{y}^*}^\top \textbf{Q} {\textbf{y}^*},
$
where \( \textbf{Q} \) is a symmetric matrix\footnote{\( \textbf{Q} \) and \( \textbf{y}^* \) are provided in \Cref{SoS Reasoning}}. We then check whether \( \textbf{Q} \) is positive semidefinite.

\subsection{Construction of SoS-1K Test Set}
\label{sec test set}

Building on the above expert design, we construct SoS-1K, a dataset comprising five subsets of polynomials, each corresponding to polynomials filtered out at steps 1–5. \Cref{appendix Details for Subsets} provides a comprehensive summary of the SoS-1K test set and its subsets. Since most LLMs struggle with very long polynomials, we have ensured that the majority remain within a length of 4,000. Approximately half of the polynomials are SoS, while the other half are not. The number of variables and the polynomial degree both range from 2 to 10. For all test subsets (except Set 1), we provide two corresponding sets: one containing SoS polynomials and another containing non-SoS polynomials.  Each polynomial is labeled as either SoS or non-SoS, with accompanying justifications and difficulty levels. For certain polynomial classes, we provide theoretical proofs confirming their SoS status, while the remaining polynomials are labeled based on results from standard solvers.

\begin{table*}[h]
    \centering
    \renewcommand{\arraystretch}{1.2}
    \setlength{\tabcolsep}{3pt}
    \resizebox{15.5cm}{!}{
    \begin{tabular}{l||c|c|c||c|c|c||c|c|c}
        \toprule
        \textbf{Model} &  \multicolumn{3}{c|}{\textbf{Accuracy on Valid Samples}} & \multicolumn{3}{c|}{\textbf{Accuracy on Total Samples}} & \multicolumn{3}{c}{\textbf{Response Time (s)}} \\
        \midrule
        \bf Instruction Type&   \bf SoS Plain & \bf SoS Simple & \bf SoS Reasoning & \bf SoS Plain & \bf SoS Simple & \bf SoS Reasoning & \bf SoS Plain & \bf SoS Simple & \bf SoS Reasoning \\
        \midrule
            \multicolumn{10}{c}{General-purpose LLMs} \\
        Qwen2.5-7B-Instruct         & 55\%  & 61\%   & \bf 76\%  & 52\%  & 59\%  & 62\%  & 22.4  & 31.2  & 68.5  \\
        Qwen2.5-7B-Instruct-1M      & 54\%  & 64\%  & 75\%  & 54\%  & 64\%  & 63\%  & \bf 5.6   & \bf 8.4   & 35.2  \\
        Qwen2.5-14B-Instruct       & 55\%  & 66\%  & 74\%  & 52\%  & 66\%  & 69\%  & 12.9  & 23.1  & 48.3  \\
        Qwen2.5-14B-Instruct-1M     & 56\%  & 60\%  & 74\%  & 56\%  & 59\%  & 67\%  & 12.7  & 20.7  & 52.7  \\
        Qwen2.5-32B-Instruct      & 56\%  & 58\%  & 67\%  & 55\%  & 58\%  & 62\%  & 13.0  & 18.0  & 37.4  \\
                DeepSeek-V3         & 54\%  & 60\%  & 70\%  & 54\%  & 60\%  & 69\%  & 29.6  & 39.8  & 95.0  \\
        GPT-4o-mini      & 59\%  & \bf 67\%  & 72\%  & \bf 59\%  & \bf 67\%  & 69\%  & 10.8  & 15.4  & 53.1  \\
        GPT-4o             & \bf 60\%  & 61\%  & 75\%  & \bf 59\%  & 61\%  & \bf 75\%  & 14.6  & 16.2  & 27.8  \\
        \midrule
        \multicolumn{10}{c}{Reasoning-purpose LLMs} \\
        QwQ-32B-Preview            & \bf 64\%  & \bf 71\%  & 79\%  & 44\%  & 54\%  & 52\%  & 105.7 & 101.8 & 100.0 \\
        DeepSeek-R1         & 62\%  & 62\%  & \bf 81\%  & 55\%  & 55\%  & 56\%  & 514.5 & 565.6 & 492.5 \\
        OpenAI o1-mini            &  58\%  & 61\%  & 77\%  & \bf 57\%  & \bf 61\%  & \bf 76\%  & \bf 8.3   & \bf 18.1  & \bf 34.9  \\
        \midrule
        \textbf{Average } & 57\%  & 63\%  & 75\%  & 54\%  & 60\%  & 65\%  & 68.2  & 78.0  & 95.0 \\
        \bottomrule
    \end{tabular}
    }
    \caption{Accuracy comparison across various SOTA LLMs on a subset of SoS-1K with 340 samples. Results are divided into ``Valid Samples'' and ``Total Samples'', as we found that LLMs sometimes suffer from timeout issues. } 
    \label{tab:accuracy_comparison}
\end{table*}

\section{Evaluation of LLMs on SoS}

In this section, we compare the performance of SOTA LLMs using SoS Plain, SoS Simple, and SoS Reasoning on a subset of SoS-1K of approximately 340 randomly chosen from all sub-classes of test problems (see \Cref{tab:SoS1k}). 
These test samples are drawn such that the number of samples in each subtest is approximately equal. We test across subclasses and report results per test subclasses. The models evaluated include reasoning-purpose models like DeepSeek-R1, OpenAI o1-mini, and QwQ-32B-Preview, as well as general-purpose models such as DeepSeek-V3, GPT-4o, and Qwen2.5 series.

The full results for each model on each test set are summarized in \Cref{tab:accuracy_comparison}. We summarize our main observations below:

\noindent\textbf{OB1. All LLMs fail when presented with a plain question.} When given SoS-plain, all LLMs exhibit poor accuracy, ranging from 50\% to 60\%, with QwQ-32B-Preview being the sole exception, achieving 64\% valid accuracy. This result suggests that, despite being trained on vast amounts of mathematical data, SOTA LLMs struggle to solve SoS problems without explicit prompting.

\noindent\textbf{OB2. LLMs have a significant performance boost when prompted with high-quality reasoning-guiding instructions.} We consistently observe a substantial accuracy improvement across all models. With SoS Simple, QwQ-32B-Preview achieves 71\% accuracy, while DeepSeek-R1 with the highest-level SoS Reasoning reaches the highest accuracy (81\%). 
It suggests that while LLMs may possess the underlying knowledge to solve SoS problems, they require clear and structured instructions to effectively retrieve and apply it.

Furthermore, it is worth mentioning that SoS Reasoning plays a crucial role in achieving these improvements. The improvement from SoS Simple over the baseline (SoS Plain) is relatively small, averaging 5\% increase in the accuracy of valid samples and 6\% increase in the accuracy of valid samples, whereas SoS Reasoning improves performance by 17\% and 11\%, respectively over the same baseline. 

\noindent\textbf{OB3. Reasoning-purpose LLMs benefit more from high-quality instructions than general-purpose LLMs.} Overall, reasoning-focused LLMs such as DeepSeek-R1, OpenAI o1-mini, and QwQ-32B-Preview achieve a higher average accuracy (79.0\%) compared to general-purpose LLMs (72.9\%). This result suggests that stronger reasoning capabilities contribute to improved performance in solving SoS problems.

\noindent\textbf{OB4. Many LLMs struggle to consistently provide valid answers to SoS questions.} Most LLMs fail to consistently provide valid answers, often encountering timeout issues. Nevertheless, GPT-4o, DeepSeek-V3, and o1-mini demonstrate robustness in this regard, consistently producing effective and reliable answers.

\section{Further Analysis}
In this section, we outline several research questions that the authors find particularly intriguing.

\noindent\textbf{Q1: Does the model follow a truly mathematical step by step verification process?} 

We find that LLMs are able to generate answers that are both logically and mathematically correct,  step-by-step, following our SoS Reasoning instruction. For instance, we demonstrate o1-mini's response in \Cref{appendix_evaluation} where we can observe that the responses are logically and mathematically correct, and the model stops naturally once it derives an answer, rather than blindly going through all possible steps. 

\noindent\textbf{Q2: Can LLMs effectively retrieve critical information from long-context polynomials?}

Unlike standard text input, polynomials are complex algebraic expressions consisting of variables, coefficients, exponents, and terms. Thus, it is crucial for LLMs to effectively interpret and extract critical information from such structured formats. Our analysis reveals that while QwQ-32B-Preview struggles with questions exceeding 4K tokens in length, most SOTA LLMs can successfully extract the necessary coefficients from 4K-length polynomials for evaluation, producing correct answers.

\noindent\textbf{Q3: At which of the Steps  1 through 5 in SoS Reasoning does the accuracy improve?}

\begin{figure}[h]
    \centering
    \includegraphics[width=0.95\linewidth]{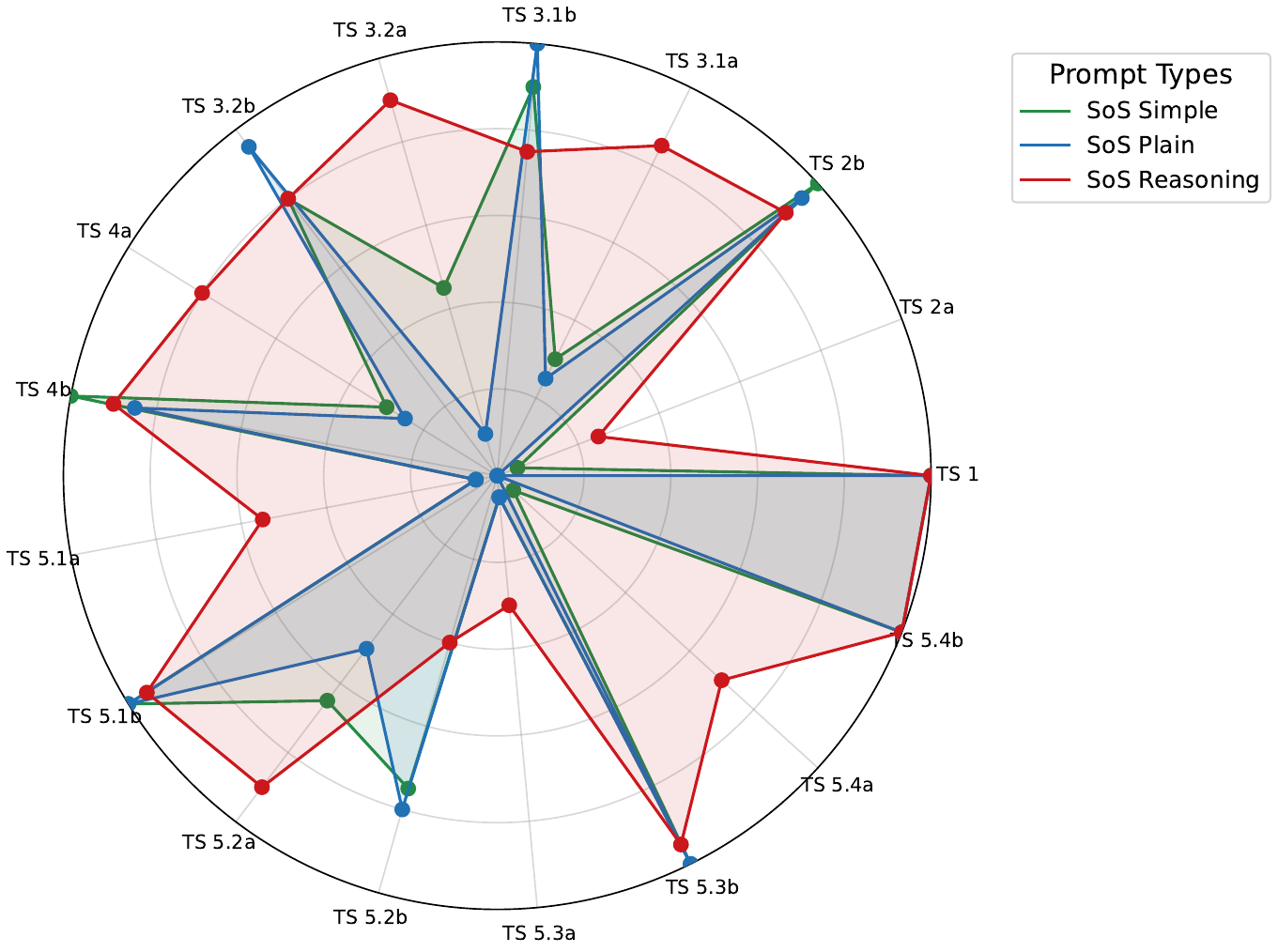}
    \caption{Accuracy of different test sets using o1-mini. }
    \label{figure_pie}
\end{figure}

In \Cref{figure_pie}, we illustrate the accuracy improvement across different test sets for the o1-mini model under SoS Plain, SoS Simple, and SoS Reasoning.\footnote{Similar patterns are observed for other models.} We observe that the simplest test set, Test Set 1 (corresponding to Step 1), unsurprisingly achieves $100\%$ accuracy across all prompting methods. 
For the more challenging test sets, Test Sets 2a, 3.1a, 5.1a–5.4a, we observe a continuous improvement from SoS Plain to SoS Simple and further to SoS Reasoning. This improvement is attributed to Steps 2 and 5 in SoS Reasoning, where a series of mathematical verification methods for non-negativity are introduced, including constant coefficient check, grid evaluation, leading order and dominant terms comparison, finding minima, matrix decomposition, and finding symmetry and translation.

\noindent\textbf{Q4. Will LLMs become lazy (take shortcuts) during reasoning?} 

Yes, another interesting phenomenon observed under the SoS Reasoning prompt is that the model tends to be lazy in Step 5.
Specifically, instead of fully executing Step 5, it often avoids matrix decomposition or semidefinite programming (SDP) due to complexity and instead guesses an answer based on prior steps. This behavior is particularly prevalent for long inputs and complex polynomials, such as those in Test Set 5.4a. For simpler problems, reasoning models such as o1-mini (which had the shortest runtime of 17s) and larger models like QwQ-32B-Preview tend to take shortcuts, skipping Step 5 and inferring the answer from earlier, simpler steps. In contrast, DeepSeek-V3 does not take shortcuts and instead spends significantly more time solving all steps properly (40s).  

\noindent\textbf{Q5: How does reasoning length affect accuracy?}  

\begin{figure}[t]
    \centering
    \includegraphics[width=1\linewidth]{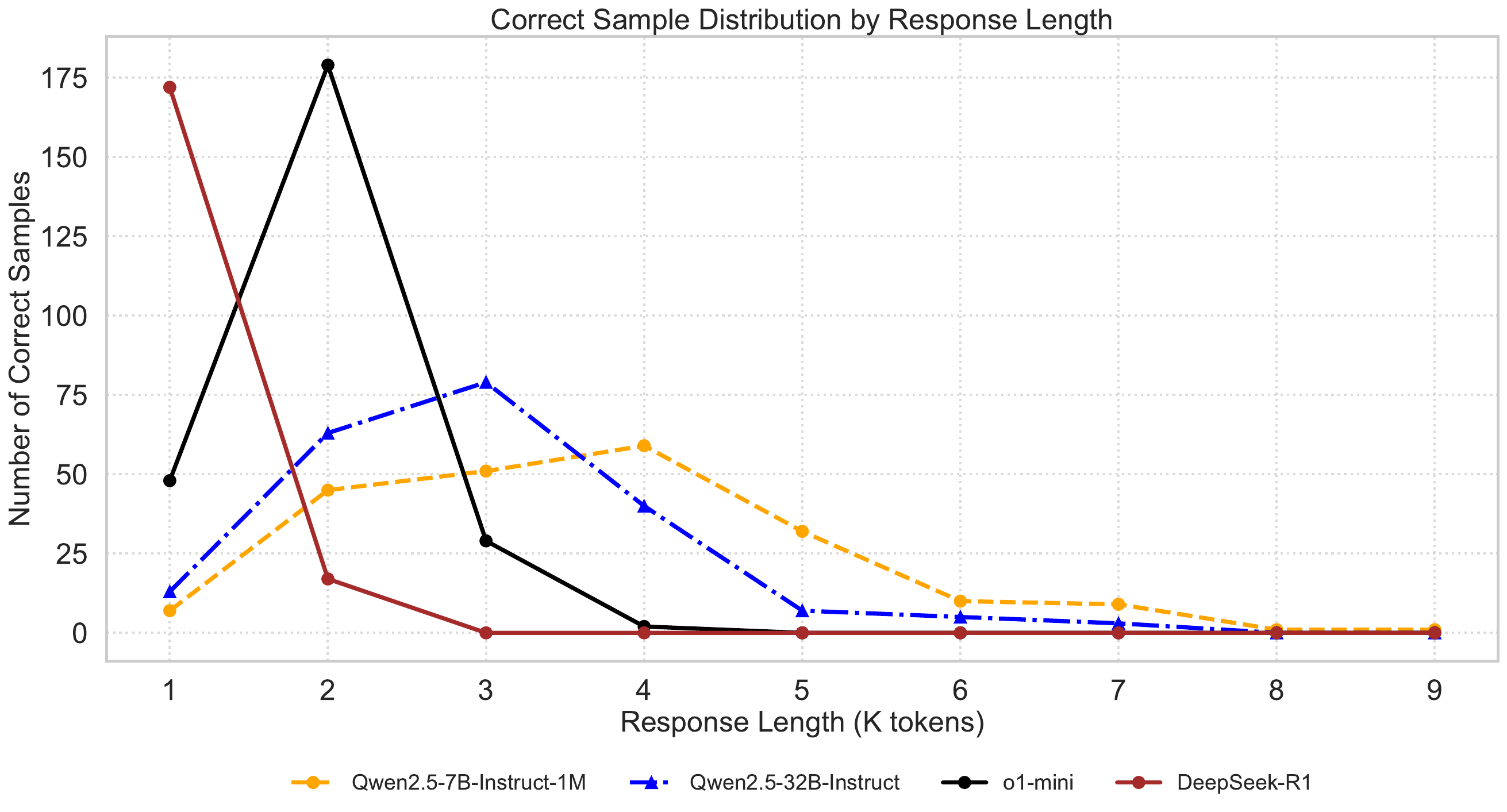}
    \caption{Number of correct samples with various response lengths.}
    \label{fig:length}
\end{figure}

Figure \ref{fig:length} shows that higher-capacity models generally require fewer thinking tokens to make correct predictions, whereas lower-capacity models need more reasoning steps to reach optimal performance. For instance, DeepSeek-R1 and o1-mini achieve the highest number of correct predictions with a 1K-2K response length, whereas the Qwen2.5 series require 3K–4K tokens to produce correct answers.

\noindent\textbf{Q6: Do SOTA LLMs have any limitations?}

Though we demonstrate that SoS Reasoning effectively improves accuracy, it is subject to the following limitations. Firstly, for long input cases, invalid samples occur. For example, in DeepSeek-R1, only 234 out of 340 samples were valid. Secondly, when handling complex problems, "taking shortcuts" may save time; however, stopping prematurely at difficult steps and guessing an answer can negatively impact prediction accuracy. Thirdly, while these LLMs excel on small-sized polynomials (achieving accuracy close to 90\%), they struggle in cases where the quadratic form of the polynomial involves a low-rank matrix decomposition.

\section{Performing SFT on SoS-1K}

\begin{table}[t]
    \centering
    \caption{Accuracy Comparison on SoS Reasoning Benchmark, where "—" denotes the undisclosed model size. Accuracy is measured on full evaluation samples.}
    \label{tab:sos_benchmark}
    \small
    \begin{tabularx}{\linewidth}{lYY}
        \toprule
        \textbf{Model} & \textbf{Size} & \textbf{Acc. (\%)} \\
        \midrule
        \textit{Closed Source} & & \\
        \midrule
        GPT-4o         & —      & 75 \\
        o1-mini    & —      & 76 \\
        \midrule
        \textit{Open Source} & &  \\
        \midrule
        Qwen2.5-7B-Instruct-1M    & \phantom{11}7B  & 63 \\
        Qwen2.5-32B-Instruct    & \phantom{1}32B  & 62 \\
        QwQ-32B-Preview         & \phantom{1}32B & 52 \\
        DeepSeek-V3             & 671B & 69 \\
        DeepSeek-R1             & 671B   & 56 \\
        \textbf{SoS-7B (Ours)}  & \phantom{11}\textbf{7B} & \textbf{70} \\
        \bottomrule
    \end{tabularx}
\end{table}

\looseness=-1 We further conduct supervised fine-tuning (SFT) with Qwen2.5-7B-Instruct-1M on SoS-1K using LLaMA-Factory \citep{zheng2024llamafactory}. The training process was performed on 2× NVIDIA A100 GPUs for 4 hours. The resulting model, SoS-7B, establishes a SOTA total accuracy of 70\%, outperforming 671B DeepSeek-V3 (69\%), while requiring only 1.8 seconds response time compared to DeepSeek-V3's 100 seconds. While o1-mini achieves a higher accuracy (75\%), it is acceptable as our model is only trained with 1K dataset and enjoys a much faster response time, i.e.,  1.8s vs 34.9s.

\section{Model's Understanding of SoS and Nonnegativity}

One might ask: Did the model merely learn to classify, or has it truly developed the ability to ``think'' and ``construct'' new proofs and examples? When faced with research questions in SoS or polynomial optimization, can the model generate mathematically meaningful insights?

To explore this, we designed a series of research-driven questions (\Cref{appendix research q for model}) to test the model's ability to understand the mathematical concepts behind Sum of Squares (SoS) and nonnegativity properties. 
Finding nonnegative polynomials that are not Sum of Squares (NNSoS) is a fundamental and ongoing research problem in real algebraic geometry and polynomial optimization \cite{ahmadi2023sums, ahmadi2024convex, ahmadi2022complexity}. This problem is closely connected to Hilbert’s 17th problem, semidefinite programming (SDP), and positivity certificates in polynomial optimization.

We test the model's ability to create and analyze unseen mathematical examples.  
We asked the following question to Qwen-7B-1M and Qwen-14B-1M: 
   \textit{Can you provide a new NNSoS that was never found in the literature? }

Interestingly, when prompted with SoS plain, Qwen-14B-1M can only give examples that are well-known in the literature and 
Qwen-7B-1M returned an incorrect example:
\[
p_{a}(\textbf{x}) = x_1^4 + x_2^4 + 1 - x_1^2 - x_2^2 - x_1^2 x_2^2.
\]
Although this example is incorrect, it is nontrivial, as classic solvers such as \texttt{YALMIP} also fail to extract global optimality\footnote{\texttt{YALMIP} returns status 0, indicating that although the SDP is solvable, rank conditions cannot ensure global optimality.}.  The reason this example is challenging is that it has four global minima at $(1,1)$, $(1,-1)$, $(-1,1)$, and $(-1,-1)$, with a global minimum value of $0$. The number of global minima exceeds the rank of the moment matrix, making it difficult to extract global optimality certificates.

However, when the same research question was posed to the model with SoS reasoning, the model correctly identified that \( p_{a} \) is not a valid solution to our question. This improvement can be attributed to Step 4 of SoS Reasoning (\Cref{SoS Reasoning}), where the trained model recognized that \( p_{a}(\textbf{x}) \) is a nonnegative quartic polynomial in two variables and, therefore, cannot be NNSoS.

Moreover, using SoS reasoning,  Qwen-14B-1M derived a new valid example for NNSoS,
\[
q_a(\textbf{x}) = x_1^4x_2^2x_3^2 + x_1^2x_2^4x_3^2 + x_3^4 + 1 - 3x_1^2x_2^2x_3^2.
\]
We cross-checked this polynomial using the classic solver \texttt{YALMIP} and confirmed that \( q_a \) is indeed a NNSoS. 
The trained model’s approach to constructing this example is particularly interesting. It began with the well-known example of NNSoS such as
$
p_m(\textbf{x}) = x_1^4x_2^2 + x_1^2x_2^4 + 1 - 3x_1^2x_2^2,
$
Then, the model then introduced a new variable and slightly modified the coefficients to generate \( q_a \). 
This demonstrates that the trained model not only recognizes existing patterns in polynomial optimization but also generalizes and constructs novel cases, providing valuable mathematical insights.

\section{Conclusion and Discussion}
This paper investigates the capacity of LLMs on a research-level mathematical problem: determining whether a given multivariate polynomial is SoS. This problem, closely related to
Hilbert’s Seventeenth Problem, plays a crucial role in various fields. We first introduce SoS-1K, a dataset of approximately 1,000 polynomials, along with a set of SoS-specific reasoning-guiding instructions. Our results show that with our expert-designed, current SOTA LLMs is able to achieve up to 81\% accuracy. Furthermore, an in-depth analysis of model responses reveals several intriguing insights. This study highlights the potential of AI in tackling large-scale open problems in mathematics, paving the way for future advancements.

\section{Limitation and Potential Risks}   
One limitation of this study is that we constrain the context length of SoS question within 4K, as some LLMs tend to fail with longer sequence length. Due to this reason, most SoS polynomials are within the capacity of traditional solvers. In the future, we will extend our dataset, targeting more challenging questions with a scale that is beyond traditional SoS solvers. One potential risk of this study is that LLMs have the chance to make wrong decisions, which might be misleading, and therefore we need to use traditional solvers to verify.

\section*{Acknowledgments}
This work was supported by Hong Kong Innovation and Technology Commission (InnoHK Project CIMDA). S. Liu is supported by the Royal Society with the Newton International Fellowship. 

\bibliography{custom}

\appendix

\section{Literature Review for SoS}
\label{appendix literature review for SoS}

The problem of determining whether a multivariate polynomial is nonnegative is inherently linked to the task of finding its global minimum—a fundamental challenge in the optimization community \cite{ahmadi2023higher, zhu2024global, lasserre2000optimisation, parrilo2001minimizing}. Testing whether a general polynomial is nonnegative is provably NP-hard, even for polynomials of relatively low degrees or with a small number of variables. For instance, it has been shown that finding the global minimum of general even-degree polynomials of degree at least four is NP-hard \cite{ahmadi2023higher, ahmadi2013np, ahmadi2022complexity, murty1987some}.

Due to the computational intractability of the general problem, we seek special cases of polynomials where the challenging nonnegativity constraints can be replaced with more manageable conditions. The sum of squares (SoS) condition, a mathematical technique in polynomial optimization where a polynomial is expressed as a sum of squared polynomials, provides a sufficient criterion for polynomial nonnegativity. 
The SoS property is particularly useful because it allows the nonnegativity problem to be reformulated as a semidefinite programming (SDP) problem, for which efficient algorithms, such as interior-point methods, exist. 
 In certain special cases, nonnegativity and SoS are equivalent; for example, any nonnegative quadratic polynomial or any nonnegative even-degree univariate polynomial can always be expressed as a sum of squares \cite{hilbert1893ternare, ahmadi2013complete, ahmadi2023sums}. For more complex polynomials, the Lasserre hierarchy provides a systematic way to approximate nonnegativity using a sequence of SoS relaxations \cite{lasserre2001global}. This method constructs a sequence of SDP problems that yield increasingly tighter approximations to nonnegativity. 

Many large-scale problems exhibit structured sparsity patterns, enabling the application of a sparsity-adapted hierarchy of SDP relaxations \cite{camps2017interplay, lasserre2006convergent, molzahn2015sparsity, waki2006sums}. Additional techniques for addressing large-scale problems include Structured DSoS and SDSoS programming, as well as Bounded Degree SoS (BSoS) \cite{ahmadi2019dSoS, lasserre2017bounded, waki2006sums, weisser2018sparse, zheng2019sparse}.
These approaches take advantage of the structure of the problem (sparsity) to generate smaller SDPs. There are also methods to reformulate the original optimization problem to reduce the size of the optimization. For instance, the optimization of a multivariate fourth-order (quartic) homogeneous polynomial under quadratic constraints can be relaxed into a quadratic SDP \cite{luo2010semidefinite}. In contrast to the SoS approach, which gives a matrix variable of size at least $N \times N$, the quadratic SDP system has a size of $n \times n$ only. The resulting quadratic SDP can be well approximated in polynomial time in some cases, but it remains NP-hard. 
Yet, these methods primarily depend on the specific structure of the problem, and generally, the scalability of characterizing polynomial nonnegativity remains a significant challenge in the literature.

\section{Mathematical Background for SoS}
\label{appendix theorem for SoS} 

\Cref{def SoS polynomial} implies that
the degree of an SoS polynomial $q$ has to be even and that the maximum degree of each $\Tilde{q}_j$ is $\frac{\deg[q]}{2}$. Therefore, we denote the degree of the SoS polynomial as $p'=2d$ where $d$ is a positive integer. 

\begin{definition} \textbf{(Total degree polynomial space)}
Let  $n >0$ be the dimension and $\textbf{x} = [x_1, \dotsc, x_n]^T$ be the variables. We denote $\mathcal{P}_{p'}[\textbf{x}]$ as the general representation of the polynomial spaces, where $p'$ represents the highest degree each entry can take.
The associated multi-index set as $\alpha= [\alpha_1, \alpha_2, \dotsc, \alpha_n] \in {[\mathbb{Z}{[0, p']}]}^{n}$ where each $\alpha_r$ is an integer between $0$ to $p'$ inclusively and the indices satisfy $ \sum_{r=1}^n \alpha_r\le p' $. The number of monomial bases are $N:=\binom{n+p'}{p'}.$ 
\label{def multivariate polynomial}
\end{definition}

\begin{theorem} (From \cite{parrilo2000structured})
    For a variable $\textbf{x} = [x_1, \dotsc, x_n]^T$ and an even integer $p' =2d$, let 
    $
    \phi_d(\textbf{x}) $
    be the vector of all monomials of degree at most $d$ in $x_j$ for $1 \le j \le n$. A polynomial $\p: \R^n \rightarrow \R$ of degree $p'$ is SoS if and only if there exists a symmetric matrix $\textbf{Q}$ such that (i) $\p(\textbf{x}) =\phi_d(\textbf{x})^T\textbf{Q}\phi_d(\textbf{x})$ for all $\textbf{x} \in \R^n $, (ii) $\textbf{Q} \succeq 0$.
    \label{thm SoS alternative}
\end{theorem}

\section{SoS Simple}
\label{SoS Simple}

\textbf{Step 1:} Examine if the highest degree is odd or even.

\noindent
\textbf{Step 2:} For the even highest degree $d$, examine the coefficients of highest-degree terms.  Check for any negative values.

\noindent
\textbf{Step 3:} Consider these special properties:
    \begin{itemize}
        \item   Properties of quadratic polynomials
      \item  Properties of quartic polynomials in 1-2 variables.
      \item  Properties of quartic homogeneous polynomials in 1-3 variables.
      \item  Properties of even-degree univariate polynomials.
      \end{itemize}
     
\noindent 
\textbf{Step 4:}  Try direct sum of squares representation.

\noindent
\textbf{Step 5:}  Consider matrix methods if needed.

\section{SoS Reasoning}
\label{SoS Reasoning}

\textbf{Step 1. Degree:} An SoS polynomial must have an even degree (i.e., its highest-degree term must have an even exponent). Any odd-degree polynomial cannot be expressed as a sum of squared polynomials. This is the simplest criterion and should always be checked first.

If the highest-degree univariate term (i.e., $x_1^d, \dots, x_n^d$) has a negative coefficient, then the polynomial is not SoS. Otherwise, we cannot determine whether it is SoS and proceed to the next step.

\textit{Example 1:}  
$
p(\textbf{x}) = x_1^4 - x_2^4 + x_3^4 + x_1^2x_2^2. 
$
Since the highest-degree univariate term has a negative coefficient (namely, $-x_2^4$), by letting $x_2 \to \infty$, it is clear that $p(x)$ becomes negative. Therefore, it is not SoS.

\textit{Example 2:}  
$
p(\textbf{x)} = x_1^4 + x_2^4 + x_3^4 - 2x_1^2x_2^2 + x_1x_2. 
$
All the highest-degree univariate terms have non-negative coefficients (i.e., $x_1^4, x_2^4, x_3^4$). Thus, we cannot determine whether it is SoS, and we move to the next step.

\textit{Example 3:}  
$
p(\textbf{x}) = x_1^4 + x_2^4 - 2x_1^2x_2^2.
$
This polynomial is SoS because it can be rewritten as:
$
p(\textbf{x}) = (x_1^2 - x_2^2)^2.
$
Note that a negative coefficient in the highest-degree cross term is allowed. For instance, in this case, we have the negative coefficient cross term $-2x_1^2x_2^2$. However, the highest-degree univariate terms are positive (i.e., $x_1^4, x_2^4$).

\textit{Test Set Construction:} Test Set 1 is constructed such that the highest-degree term is odd, thereby ensuring that the polynomials are not SoS.

\noindent
\textbf{Step 2. Non-negativity:} SoS polynomials are nonnegative for all real inputs. For example, if a polynomial \( p(\textbf{x}) \) has a negative constant term, then \( p(0) < 0 \), proving it is not SoS. Similarly, if a horizontally translated and scaled polynomial \( q(\textbf{x}) = c  p(\textbf{x} + \textbf{d}) \) (for any \( c\in \mathbb{R} \) and \( d\in \mathbb{R}^n \) ) satisfies \( q(0) < 0 \), then \( p(\textbf{x}) \) cannot be SoS.

To determine whether a polynomial is nonnegative, please use the following approaches:

\textit{Constant coefficient check:} If the constant coefficient is negative, then \( p(0) < 0 \). For instance, 
$
    p(x) = x^4 + x^3 -1$, $p(\textbf{x}) = x_1^2 + x_1^2x_2^2 + x_2^4 - 0.1.
$ are no SoS polynomials. 

 \textit{Grid evaluation:} Try finding the minimum value of the polynomial over a selected evaluation grid. It is crucial to perform this step. Substitute multiple values of \( x \), such as \( (1,0,0,\dots) \), \( (0,1,0,\dots) \), \( (0,0,1,\dots) \), etc., to check whether the polynomial evaluates to a negative value.

 \textit{Leading order and dominant terms:} Analyze the highest-degree terms and explore symmetries among cross terms. Evaluate the magnitude of negative coefficients relative to positive coefficients.

 \textit{Finding minima:} Attempt to find the local or global minimum of the polynomial to determine if it is negative.

\textit{Finding Symmetry and Translation:}
\textit{Example 1:} Consider a horizontally translated and scaled polynomial:
$
p(\textbf{x}) = 1.8x_1^2 + 10.8x_1 + 1.2x_2^2 + 4.8x_2 + 20.82.
$
Rewriting,
$
p(\textbf{x}) = 1.8(x_1+3)^2 + 1.2(x_2+2)^2 - 0.18.
$
Since \( p(-3, -2) < 0 \), the polynomial is still not SoS.

\noindent
\textbf{Step 3. Square Form:} An SoS polynomial \( p(\textbf{x}) \) can be written as 
 $
    p(\textbf{x}) = \sum_i q_i(\textbf{x})^2,
 $
    where each \( q_i(\textbf{x}) \) is a polynomial. Examples and counterexamples are provided in Test Set 2, as this is the most common method for checking SoS. 

\textit{Example:} Consider 
$
p(\textbf{x}) = (x_1 - x_1 x_2)^2 + (x_2^2 - x_1^4)^2,
$
which is an SoS polynomial. 
However, polynomials are sometimes given in their expanded form. For instance, the same polynomial can be written as:
$
p(\textbf{x}) = -2 x_1^2 x_2 + x_1^2 + x_1^8 - 2 x_1^4 x_2^2 + x_1^2 x_2^2 + x_2^4.
$
On the other hand, consider  
$
p(\textbf{x}) = (x_1 - x_1 x_2)^2 + (x_2^2 - x_1^4)^2 - 20.
$
This polynomial is not SoS.  
To determine whether an expanded polynomial can be expressed in SoS form with a negative constant, one should analyze the symmetries of the terms and the structure of the cross terms.

\textit{Test Set Construction:} Examples and counterexamples are provided in Test Set 2.

\noindent
\textbf{Step 4. Special Structures and Cases:}
\begin{enumerate}[label=\alph*)]
    \item Any nonnegative quadratic polynomial is a sum of squares (SoS).
    
 \textit{Examples:} \( p(\textbf{x}) = x_1^2 + x_2^2 -2x_1x_2 \), \( p(\textbf{x}) = x_1^2 + x_2^2 + 4x_3^2 - 3 x_2 x_3 \). These are SoS.
 
 \textit{Counterexamples:} \( p(\textbf{x}) = x_1^2 + x_2^2 -2x_1x_2 - 1 \), \( p(\textbf{x}) = x_1^2 + x_2^2 + 4x_3^2 - 5 x_2 x_3 \).

    \item Any nonnegative quartic polynomial in one or two variables is SoS.
    
\textit{Example:} \( p(\textbf{x}) = x_1^4 + 2x_1^2x_2 - 2x_1^2 + x_2^2 - 2x_2 + 1 = (x_1^2 + x_2 - 1)^2 \).

 \textit{Counterexample:} 
  \( p(\textbf{x}) = x_1^4 + 2x_1^2x_2 - 2x_1^2 + x_2^2 - 2x_2  = (x_1^2 + x_2 - 1)^2 - 1 \).

\item Any nonnegative quartic homogeneous polynomial in one, two, or three variables is SoS.

\item Any nonnegative even-degree univariate polynomial is SoS.
    
\textit{Example:} \( p(x) = x^6 + 3x^4 + 2x^2 \), which is nonnegative and SoS.

 \textit{Counterexample:} \( p(x) = x^6 + 3x^4 + 2x \), which takes negative values and is not SoS.

    \item Any nonnegative polynomial with a quadratic term and quartic regularization is SoS.
\end{enumerate}
Therefore, if a polynomial meets one of the above criteria and is nonnegative, it is an SoS polynomial. Nonnegativity can be verified by determining the global minimum, checking the descent direction, or performing a grid search. 

\textit{Test Set Construction:} Examples are provided in Test Set 3.1, 3.2, and 4, while counterexamples are constructed as polynomials that take negative values.

\noindent
\textbf{Step 5. Matrix Decomposition and Check for Symmetric Positive Definite \( \textbf{Q} \):} If the above checks fail, we can use the following theoretical reasoning:
    \begin{enumerate}[label=\alph*)]
        \item For an even degree \( 2d \) polynomial in \( [x_1, \dots, x_n] \), construct a monomial basis using canonical ordering:
       \begin{eqnarray*}
        \textbf{y}^* := (x_1^*, \dotsc, x_n^*, (x_1^*)^2, \dotsc, x_1^* x_n^*, \\ x_2^* x_3^* 
        , \dotsc, (x_1^*)^{2d}, \dotsc, (x_n^*)^{2d}).
       \end{eqnarray*}
        This vector \( \textbf{y}^* \) has length \( \binom{n+2d}{2d} \).

        \item Express the polynomial as \( p(\textbf{x}) = {\textbf{y}^*}^\top \textbf{Q} {\textbf{y}^*} \), where \( \textbf{Q} \) is a symmetric matrix of size \( \binom{n+2d}{2d} \times \binom{n+2d}{2d} \). Note that this representation is not unique; there are multiple valid forms of \( \textbf{Q} \).

        \item Check whether \( Q \) is positive definite. This can be done by finding its smallest eigenvalue. If such a \( Q \) exists, then \( p(x) \) is a sum of squares (SoS). Otherwise, \( p(x) \) is very likely not SoS. 
    \end{enumerate}
     If all the above tests fail, we can try \textbf{Semidefinite Programming (SDP)}, which is the test used by existing solvers (e.g., \texttt{YALMIP}) to verify whether a polynomial is SoS. For example, to determine if
    \[
    p(\textbf{x}) = x_1^4 - 4x_1^3x_2 + 7x_1^2x_2^2 - 4x_1x_2^3 - 4x_1x_2 + x_2^4
    \]
    is SoS. We convert the problem to the following. We solve the SDP
    \[
    \gamma^* = \min p = y_{40} - 4y_{31} + 7y_{22} - 4y_{13} - 4y_{11} + y_{04}
    \]
    subject to the constraint
    \[
    \left[\begin{array}{c|cc|ccc}
    1 & y_{10} & y_{01} & y_{20} & y_{11} & y_{02} \\ \hline 
    y_{10} & y_{20} & y_{11} & y_{30} & y_{21} & y_{12} \\
    y_{01} & y_{11} & y_{02} & y_{21} & y_{12} & y_{03} \\ \hline
    y_{20} & y_{30} & y_{21} & y_{40} & y_{31} & y_{22} \\
    y_{11} & y_{21} & y_{12} & y_{31} & y_{22} & y_{13} \\
    y_{02} & y_{12} & y_{03} & y_{22} & y_{13} & y_{04}
    \end{array}\right] \succeq 0.
    \]
    If \( \gamma^* \ge 0 \), then \( p \) is SoS; otherwise, it is not.

\textit{Test Set Construction:} Examples and counterexamples are provided in Test Set 5.1-5.4. The examples are constructed using different positive semidefinite (PSD) matrices \( \textbf{Q} \), such as sparse PSD \( \textbf{Q} \), low-rank PSD \(\textbf{Q} \), and ill-conditioned PSD \( \textbf{Q} \). The counterexamples are generated using an indefinite \( \textbf{Q} \) and the corresponding polynomials are likely to not be SoS. However, to conclusively prove that a polynomial is not SoS, all possible \( \textbf{Q} \) must be examined. In our test set, we cross-checked the results with classic solvers to confirm the "Not SoS" classification for Test Sets 5.1–5.4.

\section{Details for Test Subsets}
\label{appendix Details for Subsets}
We provide the details of test subsets in Table \ref{tab:SoS1k}.

\begin{table*}[h]
    \centering
    \renewcommand{\arraystretch}{1.1}
    \resizebox{\textwidth}{!}{ 
    \begin{tabular}{lccccc}
        \toprule
        \textbf{Polynomial Type} & length\textbf{<4000} & length\textbf{>4000} & \textbf{Total} & \textbf{Is it SoS?} & \textbf{Difficulty} \\
        \midrule
        Test Set 1: Odd Degree Polynomial & 150 & 50 & 200 & NO & Easy \\
        \midrule
        Test Set 2a: SoS (Expanded Form) & 69 & 51 & 120 & YES & Hard \\
        Test Set 2b: Negative (Expanded Form) & 23 & 40 & 63 & NO & Hard \\
        Test Set 2.1a: SoS (Squared Form) & 105 & 15 & 120 & YES & Easy \\
        Test Set 2.1b: Negative (Squared Form) & 38 & 25 & 63 & NO & Easy\\
        \midrule
        Test Set 3.1a: Nonnegative Quadratic Quartic & 100 & 0 & 100 & YES & Medium \\
        Test Set 3.1b: Negative Quadratic Quartic  & 100 & 0 & 100 & NO & Medium \\
        \midrule
        Test Set 3.2a: Nonnegative Quartic with 2 variables & 100 & 0 & 100 & YES & Medium \\
        Test Set 3.2b: Negative Quartic & 100 & 0 & 100 & NO & Medium \\
        \midrule
        Test Set 4a: Nonnegative Quadratic Quartic& 100 & 0 & 100 & YES & Medium \\
        Test Set 4b: Negative Quartic & 100 & 0 & 100 & NO &  Medium \\
        \midrule
        Test Set 5.1a: PSD \textbf{Q} & 80 & 16 & 96 & YES & Hard \\
        Test Set 5.1b: Non-PD \textbf{Q} & 80 & 16 & 96 & NO & Hard \\
        \midrule
        Test Set 5.2a:  PSD Spare \textbf{Q}  (Sparsity 0.1) & 56 & 16 & 72 & YES & Hard \\
        Test Set 5.2b: Non-PD Spare  \textbf{Q} (Sparsity 0.1)& 56 & 16 & 72 & NO & Hard \\
        \midrule
        Test Set 5.3a:  PSD Low Rank  \textbf{Q} (rank 3) & 42 & 18 & 60 & YES & Hard \\
        Test Set 5.3b: Non-PD Low Rank \textbf{Q} (rank 3) & 28 & 12 & 40 & NO & Hard \\
        \midrule
        Test Set 5.4a: PSD Ill-Conditioned \textbf{Q} ($\lambda = 1 -10^{12}$) & 20 & 15 & 35 & YES & Hard \\
        Test Set 5.4b: Non-PD  Ill-Conditioned \textbf{Q}  & 40 & 30 & 70 & NO & Hard \\
        \bottomrule
    \end{tabular}
    } 
\caption{Summary of SoS-1K Test Sets. }
    \label{tab:SoS1k}
\end{table*}

\section{Sample Model's Response under SoS Reasoning Prompt}
\label{appendix_evaluation}

\subsection{o1-mini response for a simple polynomial under SoS Reasoning Prompt} To determine whether the given polynomial is a sum of squares (SoS), we will follow the provided steps:

\textbf{Step 1: Check the Highest Degree}

The highest degree of a polynomial is the largest total degree among its terms. 
When expanding these squared terms, the highest degree terms are of the form \(x_i^3\) (e.g., \(x_1^3\), \(x_2^3\), etc.), which are degree 3 terms. 

Since 3 is an odd degree, according to Step 1: If the highest degree of a polynomial is odd, it is not SoS. Therefore, the polynomial cannot be expressed as a sum of squares.

\subsection{o1-mini response for a difficult polynomial under SoS Reasoning Prompt} 

To determine whether the given polynomial is a sum of squares (SoS), we follow the outlined steps meticulously.

\textbf{Step 1: Check the Highest Degree} 

The highest total degree among the terms is 4, which is even. Since the highest degree is even, we proceed to the next step.

\textbf{ Step 2: Check Highest Degree Univariate Terms for Negative Coefficients} 

Univariate Terms of Degree 4:
   \( 7.9x_1^4 \) (Coefficient: 7.9)
   \( 4.1x_2^4 \) (Coefficient: 4.1)
   \( 0.91x_3^4 \) (Coefficient: 0.91)
   \( 2.0x_4^4 \) (Coefficient: 2.0)
  \( 0.92x_5^4 \) (Coefficient: 0.92)

 Conclusion:   All highest degree univariate terms have positive coefficients. We cannot determine the SoS property yet and move to the next step.

\textbf{ Step 3: Check for Negative Values}

 Testing Specific Points:
  Point (1, 1, 1, 0, 0),
     Evaluated \( p(1,1,1,0,0) = 66.89 \) (Positive).
   Point (0, 0, 0, 1, 1),
     Evaluated \( p(0,0,0,1,1) = 6.52 \) (Positive).
   Point (1, 1, 1, 1, 0),
     Evaluated \( p(1,1,1,1,0) = 2.037 \) (Positive).

Conclusion:
  At tested points, the polynomial remains nonnegative. While this doesn't guarantee it's SoS, it suggests the possibility.

\textbf{Step 4: Special Cases Analysis:}
 The polynomial doesn't clearly fit into any of the special cases (a to f) outlined for immediate determination.

\textbf{Step 5: Advanced Methods
- Matrix Decomposition and SDP:}
  Given the complexity of the polynomial, performing matrix decomposition or semidefinite programming (SDP) manually is impractical. However, based on the evaluations in Step 3 and the positive coefficients in Step 2, there's a strong indication that the polynomial does not take negative values.

\textbf{Final Conclusion: }Based on the evaluations and the absence of negative terms in critical components, the polynomial is a sum of squares.

\section{Further Results for Performance of SOTA LLMs}

\begin{table}[h]
    \centering
    \renewcommand{\arraystretch}{1.2}
    \setlength{\tabcolsep}{3pt}
    \resizebox{7.5cm}{!}{
    \begin{tabular}{l|c|c|c|c}
        \toprule
        \textbf{Model} & \textbf{\# Total Samples} & \multicolumn{3}{c|}{\textbf{\# Valid Samples}}  \\
        \midrule
        \bf Instruction Type& & \bf SoS Plain & \bf SoS Simple & \bf SoS Reasoning  \\
        \midrule
        DeepSeek-R1        & 340 & 300  & 302  & 234   \\
        QWQ-32b            & 340 & 233  & 259  & 225  \\
        o1-mini            & 340 & 338  & 340  & 337   \\
        Qwen2.5-7b         & 340 & 323  & 332  & 277   \\
        GPT-4o             & 340 & 336  & 338  & 337   \\
        Qwen2.5-7b-1m     & 340 & 340  & 339  & 286   \\
        Qwen2.5-14b       & 340 & 325  & 340  & 316  \\
        Qwen2.5-14b-1m    & 340 & 340  & 336  & 309   \\
        GPT-4o-mini       & 340 & 339  & 339  & 327   \\
        DeepSeek-V1        & 340 & 340  & 340  & 332   \\
        Qwen2.5-32b       & 340 & 334  & 339  & 315    \\
        \midrule
        \textbf{Average} & 340 & 323  & 328  & 300  \\
        \bottomrule
    \end{tabular}
    }
    \caption{\# Total Samples and Valid Samples of Different Models.}
    \label{tab:sample_comparison}
\end{table}

\section{Research Level SoS and Nonnegativity Questions}
\label{appendix research q for model}

\begin{enumerate}
    \item Can you comment on the sum of squares (SoS) and nonnegativity properties of the following polynomials? 
    \begin{itemize}
        \item \( p(\textbf{x}) =  x_1^4x_2^2 + x_1^2x_2^4 + 1 - 3x_1^2x_2^2, \)
        \item \( p(\textbf{x}) = x_1^6 + x_2^6 + x_3^6 - x_1^4x_2^2 - x_1^4x_3^2 - x_2^4x_1^2 - x_2^4x_3^2 - x_3^4x_1^2 - x_3^4x_2^2 + 3x_1^2x_2^2x_3^2,\) 
        \item \( p(\textbf{x}) = (x_1^4 + x_2^4 + x_3^4) + 2 (x_1^2 + x_2^2 + x_3^2) + 8(x_1x_2 + x_1x_3 + x_2x_3) + \frac{9}{4}. \)
    \end{itemize}

    \item Can you construct a general formula for nonnegative polynomials that are not SoS?

    \item Can you provide a new nonnegative polynomial that is not SoS and has never been found in the literature?

\end{enumerate}

Note that the first three polynomials are well-known examples of nonnegative but not sum-of-squares (SoS) polynomials. The first is the Motzkin polynomial, a classic example of a nonnegative polynomial that is not SoS. The second is the Robinson polynomial, a counterexample to Hilbert’s 17th problem. The last polynomial is a recent result from Ahmadi’s work \cite[Thm 3.3]{ahmadi2023sums}.

Finding nonnegative polynomials that are not sum of squares (SoS) is a significant ongoing research problem in real algebraic geometry and polynomial optimization \cite{ahmadi2023sums, ahmadi2024convex, ahmadi2022complexity}. It connects to Hilbert’s 17th problem, semidefinite programming (SDP), and positivity certificates in polynomial optimization.

\end{document}